\documentclass[letterpaper, 10 pt, conference]{ieeeconf}  

\newcommand{\method}{GAM} 
\newcommand{\methodlong}{Grounded Action Model}

\newcommand{\rev}[1]{#1}
\IEEEoverridecommandlockouts                              
\usepackage[utf8]{inputenc}
\usepackage{textgreek}
\usepackage{graphicx}
\usepackage{amsmath}
\usepackage{amssymb}
\usepackage{graphicx}
\usepackage{bm}
\usepackage{booktabs} 
\usepackage{multirow} 
\usepackage{xcolor}
\usepackage{caption}
\usepackage{booktabs,multirow,tabularx,xcolor}
\usepackage{array} 
\usepackage{url}                     
\usepackage{xcolor}

\title{\LARGE \bf
Grounded Action Model: 3D Grounding as a Foundation for Robotics
}

\author{
    Gehao Zhang$^{1}$, Weikai Huang$^{2}$, Shailesh Shailesh$^{3}$, Yiyan Peng$^{1}$,
    Jiafei Duan$^{3,\dagger}$, Ranjay Krishna$^{2,\dagger}$\\
    {\small $^{1}$Northwestern University \quad
            $^{2}$University of Washington \quad
            $^{3}$National University of Singapore \quad}\\
    {\small $^{\dagger}$Equal advising.}\\
        {\small \textcolor{red!60}{\url{https://grounded-action-model.github.io/}}}
}

\begin{document}

\maketitle
\thispagestyle{empty}
\pagestyle{empty}

\begin{abstract}
Manipulation policies must know which objects matter and where they are, yet the pretrained backbones that current robot foundation models build on, from language in vision-language-action models (VLAs) to video generation in world-action models (WAMs), do not directly require this metric grounding, leaving it to be learned implicitly from robot demonstrations. We propose \textit{Grounded Action Models} (GAMs), a new paradigm of robot foundation models built with 3D grounding. 
GAM can be conditioned using language, points, or box prompts, which are first transformed into a shared object-centric representation of the selected objects. 
This representation captures target-focused visual features and metric object geometry, which is mixed with robot state history through a multi-stream transformer to predict action chunks. 
Although GAMs can be run autonomously, they can also serve as a low-level controller that a high-level planner controls using its various input modalities, allowing for long-horizon and memory-dependent manipulation.
On RoboTwin 2.0, GAM achieves an average success rate of 55.3\% across 50 tasks \rev{(vs. 52.0\% for Spatial Forcing)}, including 47.6\% under scene randomization \rev{(vs. 30.4\% for Abot-M0)}, with its action policy trained only on clean-scene demonstrations. On LIBERO-PRO, it achieves \rev{a state-of-the-art average success rate of 61\% (vs. 53\% for $\pi_{0.5}$)} across 16 perturbation settings, with the largest gains when targets are relocated or newly designated. 
On two real robots, \method{} retains 17/20 successes under visual shift on a bimanual YAM versus 4/20 for $\pi_{0.5}$, while its composition with a Molmo2 planner on a Franka achieves 64.7\% ID and 49.8\% OOD step completion on long-horizon and memory-dependent tasks.
\end{abstract}

\section{Introduction}

\begin{figure*}[t]
    \centering
    \includegraphics[width=\textwidth]{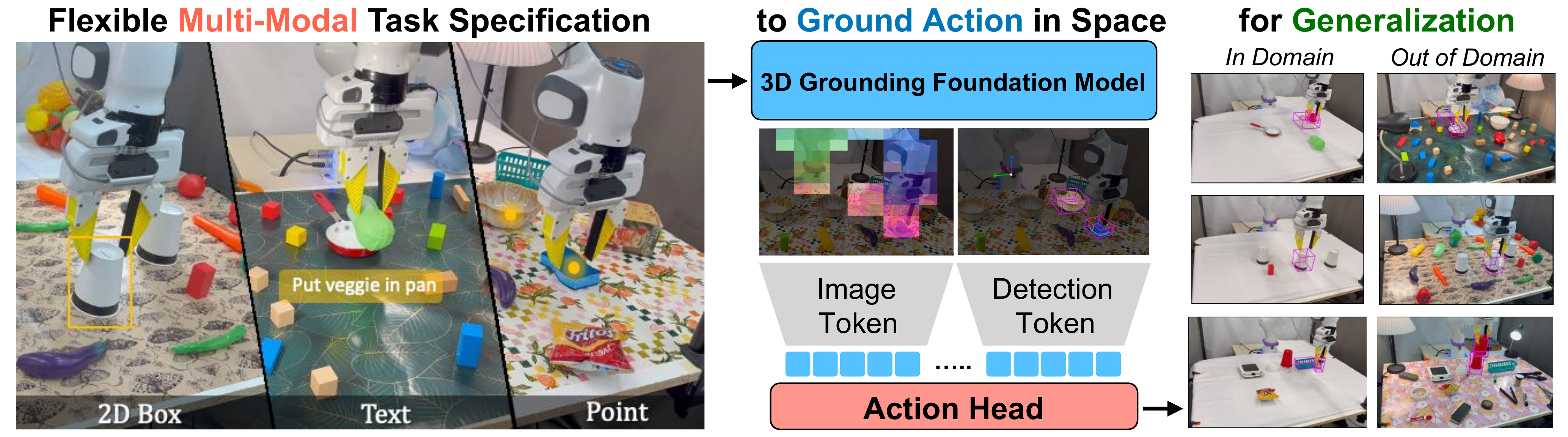}
    \vspace{-20pt}
    \caption{\textbf{3D grounding as a foundation for robot action learning.}
    \textbf{Left:} Inputs include natural-language instructions, 2D points, or 2D bounding boxes.
    \textbf{Center:} A pretrained 3D grounding foundation model connects these prompts to object-centric visual features and metric geometry, which are encoded into image and detection tokens for action prediction.
    \textbf{Right:} The resulting policy performs manipulation across in-domain and out-of-domain environments with variations in objects, backgrounds, and scene configurations.}
    \vspace{-20pt}
    \label{fig:motivation}
\end{figure*}

The choice of pretrained backbone shapes what a robot foundation model knows before it learns to act.
Two influential directions build on foundations developed for language and video generation.
Vision-language-action models (VLAs)~\cite{rt2,openvla,pi05,molmoact2} adapt pretrained vision-language backbones to robot control, transferring semantic knowledge acquired from large-scale image--text data.
Meanwhile, world-action models (WAMs)~\cite{dreamzero,fastwam,zhu2025unifiedworldmodelscoupling,xwam} build on pretrained video-generation backbones, transferring spatiotemporal priors and learning to couple predicted visual futures with robot actions.
They also raise a fundamental question:
\emph{what should a foundation model learn during pretraining to best support generalizable action learning?}

Language and video pretraining provide valuable knowledge, but their objectives do not directly require the explicit metric grounding needed for manipulation ~\cite{Spelke2007CoreK}.
A language-generation objective can reward identifying an object and describing its relationships without requiring its precise 3D location or extent.
A video-generation objective captures how scenes evolve, but favors appearance, texture, and background variation that may be irrelevant to the intended interaction.
Neither objective alone guarantees a representation that explicitly identifies the task-relevant objects and exposes their geometry for control.
The action learner must therefore connect these pretrained representations to spatially precise motor behavior using robot demonstrations.
When those demonstrations cover limited environments and layouts, the learned connection can depend on visual correlations that fail when objects move, backgrounds change, or a different target is selected\rev{~\cite{liberopro,liberoplus,robotwin2}}.

We propose \textbf{3D grounding as a foundation for robot action learning} and introduce \methodlong{} (\method{}) to instantiate this paradigm.
\method{} builds its policy on a pretrained promptable 3D grounding model~\cite{wilddet3d}, without relying on a language-generation or video-generation backbone.
Its foundation is trained to connect visual observations and task prompts to objects in metric 3D space.
This supplies the action learner with an explicit representation of \emph{which objects matter}, \emph{where they are}, and \emph{what geometry they occupy}.
Our central hypothesis is that a foundation pretrained to expose this structure provides a more direct starting point for learning manipulation.
Robot demonstrations can then teach the policy how to act on grounded objects, while the pretrained backbone supplies the object localization and spatial understanding on which those actions depend.

This choice of foundation also changes how observations are presented to the policy.
\method{} constructs two complementary streams from the grounded task objects.
Image tokens retain visual features of the selected objects and robot arm while suppressing the rest of the scene.
Detection tokens encode object point clouds together with their metric 3D positions and extents.
Together, these streams preserve both appearance information and explicit geometry while reducing exposure to irrelevant scene variation.
When a target is relocated or a different object is designated, the grounded observation follows that selection, providing the policy with updated spatial information for action prediction.
A multi-stream transformer (MM-DiT)~\cite{sd3} combines these representations with robot state history to predict chunks of absolute joint-position targets.
The grounding backbone remains frozen during policy training.

A further consequence of grounding-based action learning is a unified interface for task specification.
The backbone accepts language, 2D points, and 2D bounding boxes.
Language therefore becomes one way to specify a target, while visual prompts provide direct control over which object the policy should manipulate.
This is particularly useful when multiple objects share the same description or when coupled with a high-level planner that has already identified the intended target.
A VLM planner~\cite{molmo2} can communicate its selection directly through points, which the grounding backbone converts into spatial observations for the action policy.
For multi-step tasks, the planner updates these selections as execution progresses.
Its observation history can also support target selection when object identities or desired locations are no longer recoverable from the current image. Thus, GAM supports composition with semantic reasoning and memory while grounding action prediction in explicit object geometry.

We evaluate GAM on two simulation benchmarks and two real robots.
On RoboTwin 2.0~\cite{robotwin2}, it achieves 55.3\% average success over 50 tasks \rev{(vs. 52.0\% for Spatial Forcing)}, including 47.6\% under scene randomization \rev{(vs. 30.4\% for Abot-M0)}, despite training the action policy only on clean-scene demonstrations.
On LIBERO-PRO~\cite{liberopro}, it achieves the highest average across 16 perturbation settings (0.61 versus 0.53 for $\pi_{0.5}$~\cite{pi05}), with the largest gains when targets are relocated or newly designated.
On a bimanual YAM, \method{} and $\pi_{0.5}$ both succeed on 19 of 20 trials in distribution, but under visual shift \method{} retains 17 of 20 successes while $\pi_{0.5}$ falls to 4 of 20.
On a Franka, \method{} composes with a Molmo2 ~\cite{molmo2} planner through its point interface, achieving 64.7\% in-distribution and 49.8\% out-of-distribution step completion across two long-horizon and two memory-dependent tasks.
The corresponding out-of-distribution results are 24.0\% for $\pi_{0.5}$ and 17.1\% for MolmoAct2~\cite{molmoact2}.

These results support 3D grounding as a promising foundation for manipulation policies that must remain responsive to task-relevant spatial changes while tolerating visual variation. 
While we instantiate GAMs with WildDet3D as the backbone grounding model, trained originally for 3D box detection, we expect the community will consider alternative grounding options within this new paradigm.


\section{Related Work}

\begin{figure*}[t]
    \centering
    \includegraphics[width=\textwidth]{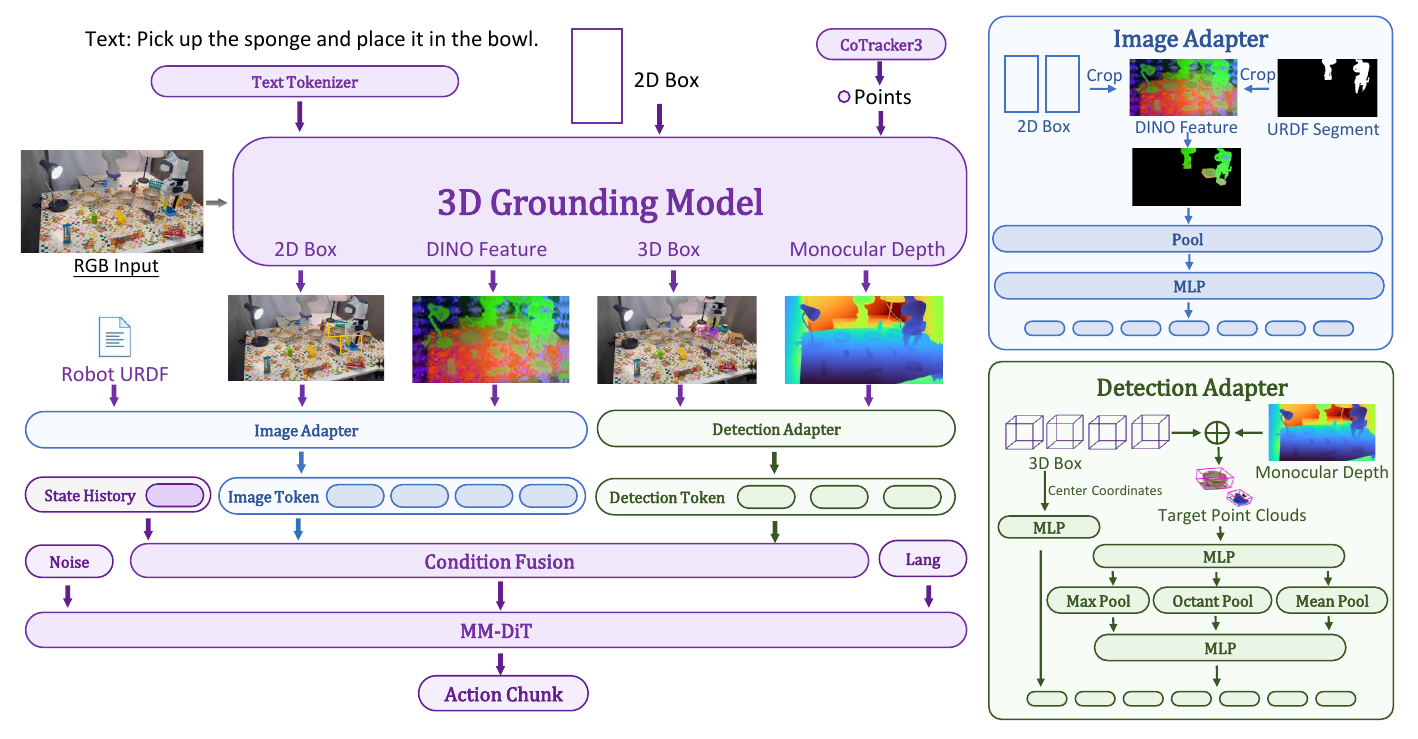}
    \vspace{-15pt}
    \caption{\textbf{\method{} architecture.} A task specified by language, 2D points, or 2D boxes is resolved by the frozen 3D grounding backbone into per-object 2D boxes, 3D boxes, and a metric depth map. From these, \method{} builds an object-centric observation: image tokens restricted to the task objects and the robot arm, detection tokens encoding each object's point cloud and pose, and a state-history token. A multi-stream transformer (MM-DiT) conditions on these tokens and a language embedding injected through adaptive layer normalization to predict a chunk of joint-position targets by flow matching.}
    \vspace{-15pt}
    \label{fig:method}
\end{figure*}

\textbf{Pretrained foundations for manipulation.} Pretrained backbones supply semantic and predictive knowledge for learning manipulation beyond individual tasks. Vision-language-action models (VLAs)~\cite{rt2,openvla,pi0,pi05,molmoact2} adapt pretrained vision-language representations to action prediction through robot demonstrations. Developments include efficient action tokenization~\cite{fast}, flow-based action generation~\cite{pi0,flowmatching}, heterogeneous co-training~\cite{pi05,knowledgeinsulation}, and intermediate visual or embodied reasoning~\cite{tracevla,cotvla,ecot,molmoact}. Another direction couples video prediction with action generation, using future visual representations and spatiotemporal priors to support control~\cite{zhu2025unifiedworldmodelscoupling,dreamzero,fastwam,imagewam}. These directions establish the value of transferring knowledge from large-scale pretraining, but semantic understanding and visual prediction do not inherently expose the selected task objects in metric 3D space. \method{} instead investigates promptable 3D grounding as the pretrained foundation for action learning~\cite{wilddet3d}. Its backbone supplies object localization, visual features, and metric geometry, allowing robot demonstrations to teach actions conditioned on this explicit spatial structure. The grounding backbone remains frozen during policy training, separating the acquisition of grounding capabilities from learning how to act on grounded objects.

\textbf{3D and grounding for manipulation.} Prior work incorporates explicit spatial structure into manipulation through voxel-based representations and multi-view projections of point clouds~\cite{peract,rvt}, or geometry-aware representations within VLAs~\cite{spatialvla}. Object-aware regularization and selective visual representations also investigate how emphasizing relevant scene content improves policy learning and generalization~\cite{oreo,selectivevisual}. Complementary approaches connect semantic reasoning to spatial control through language-conditioned 3D value maps~\cite{voxposer}, spatial affordance prediction~\cite{robopoint}, and visual traces or trajectory guidance~\cite{tracevla,hamster,hamster3d,molmoact}. These studies motivate explicit geometry, selective perception, and spatial interfaces for manipulation. GAM unifies these elements through a single promptable 3D grounding backbone that resolves language, points, and boxes into one grounded observation.
\section{GAM: Grounded Action Model}
\label{sec:method}

\method{} combines a pretrained 3D grounding backbone with an action head. Given a task prompt and the current image, the frozen backbone identifies and localizes the task objects. The action head uses their visual and geometric representations, robot state history, and a language embedding to predict a chunk of joint-position targets (Fig.~\ref{fig:method}).

\subsection{From prompt to detections}
\label{sec:method_backbone}

At time $t$, the grounding backbone receives an RGB image $o_t$ and a task specification $\ell$ provided as language, 2D points, or 2D boxes, identifying the $N$ objects involved in the task. For language input, a span-tagging head trained on synthesized instruction templates over a frozen Flan-T5 encoder~\cite{t5} extracts the task-relevant object phrases; for example, ``pick up the sponge and place it in the bowl'' yields \emph{sponge} and \emph{bowl}, which are used as separate grounding queries. Point and box inputs specify the objects directly. The resulting queries are passed to WildDet3D~\cite{wilddet3d}, the promptable 3D grounding backbone $\mathcal{G}$ adopted in this work. For each queried object $i$, $\mathcal{G}$ produces a 2D detection box $u_i$ and a metric 3D box $b_i$. It also predicts a dense metric depth map $\hat{D}_t$ from the RGB image and exposes the dense features $F_t=\mathrm{Enc}_{\mathcal{G}}(o_t)$ of its visual backbone.

All three prompt modalities are mapped to a common object-centric representation used to construct the policy observation (Sec.~\ref{sec:method_obs}). This unified interface allows a VLM planner~\cite{molmo2} to specify target objects directly through 2D points, while a separate language condition specifies the operation to perform (Sec.~\ref{sec:method_head}).

\subsection{From detections to observation}
\label{sec:method_obs}

\method{} constructs image and detection tokens from $\{(u_i,b_i)\}_{i=1}^{N}$, $F_t$, and $\hat{D}_t$ to represent task-relevant visual information and object geometry. These tokens are combined with a state-history token representing the robot's recent joint states to condition action prediction.

\textbf{Image tokens.} We average-pool $F_t$ to a $G\times G$ grid with $G=16$ and retain cells that overlap either the 2D detection boxes $\{u_i\}$ of the task objects or the robot-arm silhouette. The silhouette is obtained by posing the robot's URDF meshes using forward kinematics and projecting them into the camera with the known calibration. A cell is retained if some $u_i$ covers at least 30\% of its area or the robot silhouette covers more than 10\%. Features in the remaining cells are set to zero, and a shared linear adapter maps the grid into $G^2$ image tokens. This representation preserves visual information about the selected objects and their spatial relationship to the gripper while suppressing direct access to features outside these regions.

\textbf{Detection tokens.} We construct each object's metric geometry from the predicted depth $\hat{D}_t$ and the detector's oriented 3D box $b_i$. The full depth map is back-projected using the camera intrinsics and transformed into the robot base frame, yielding a scene point cloud; the points inside a slightly inflated $b_i$ are cropped out as the object point cloud, and $P=512$ of them are sampled to form $P_i$. The box provides the object's center $c_i$, extents $e_i$, and rotation $r_i$ in the base frame. A point-cloud encoder $\phi$ applies a shared per-point MLP and aggregates the point features by global max-pooling, global mean-pooling, and max-pooling within each of eight octants around the centroid, producing a 512-dimensional shape feature $\phi(P_i)$. The object descriptor is
\begin{equation}
z_i=\big[\,\tilde u_i,\;c_i,\;\gamma(c_i),\;\phi(P_i),\;e_i,\;r_i\,\big],
\label{eq:det}
\end{equation}
where $\tilde u_i$ is the normalized 2D box, $\gamma(c_i)$ contains Fourier features of the center with eight frequencies, and $r_i$ is the 6D representation of the box rotation. A linear adapter maps $z_i$ into $M=8$ detection tokens of width $d$, yielding $N \times M$ detection tokens for $N$ task objects. Undetected slots use a learned null token in all eight positions, preserving the token budget. These tokens expose the estimated shape, location, extent, and orientation of each selected object directly to the action policy.

\textbf{State history token and fusion.} We concatenate the robot's joint positions at the current and previous frames and project them into a single state-history token, which supplies the arm's current configuration and its most recent motion. Image, detection, and state-history tokens receive learned modality embeddings and pass through a shared self-attention layer for condition fusion. The resulting sequence $C\in\mathbb{R}^{L\times d}$, with $L=G^2+MN+1$, conditions the action transformer. The fused sequence preserves the modality-specific token positions, allowing the action transformer to process image, detection, and state-history tokens as separate streams.

\subsection{From observation to actions}
\label{sec:method_head}

The action head uses a multi-stream transformer (MM-DiT)~\cite{sd3} $v_\theta$ with 12 blocks that predicts a chunk $A_t=(a_t,\dots,a_{t+H-1})$ of absolute joint-position targets together with gripper commands. It operates on four streams: the image, detection, and state-history segments of $C$, and the noisy chunk $A_t^\tau$ embedded as $H$ action tokens. Action and condition tokens each carry learned positional embeddings. Each block uses separate query, key, value, output-projection, feed-forward, and adaptive layer-norm parameters for each stream, with timestep and language conditioning injected through adaptive layer normalization. Joint attention over the concatenated tokens enables information exchange among all four streams. An RMSNorm followed by a small MLP maps the action tokens to the predicted velocity.

\textbf{Language conditioning.} Let $e_s\in\mathbb{R}^{768}$ denote the sentence embedding of the instruction $s$. We add a learned projection of $e_s$ to the flow-timestep embedding:
\begin{equation}
h_{\tau,s}=\mathrm{TimeEmbed}(\tau)+W_{\ell}e_s+b_{\ell},
\label{eq:lang_cond}
\end{equation}
where $W_{\ell}\in\mathbb{R}^{576\times768}$ and $b_{\ell}\in\mathbb{R}^{576}$. The resulting vector conditions the adaptive layer normalization in each transformer block, providing operation semantics alongside the grounded object representations. Both $W_{\ell}$ and $b_{\ell}$ are initialized to zero, preserving the original model at initialization. This conditioning introduces no additional tokens.

During policy training, we optimize only the action head, comprising the image adapter, the detection adapter with its point-cloud encoder $\phi$, the condition fusion layer, the language projection, and the multi-stream action transformer, while keeping $\mathcal{G}$ frozen. We train $v_\theta$ with flow matching~\cite{flowmatching, pi0}: for a demonstration chunk $A_t$, noise $\epsilon\sim\mathcal{N}(0,I)$, and $\tau\in[0,1]$, we construct $A_t^\tau=\tau A_t+(1-\tau)\epsilon$ and train the model to predict the velocity $A_t-\epsilon$:
\begin{equation}
\mathcal{L}=\mathbb{E}_{A_t,\epsilon,\tau}\left[\left\|v_\theta(A_t^\tau\mid\tau,C,e_s)-(A_t-\epsilon)\right\|_2^2\right].
\label{eq:fm}
\end{equation}
At inference, $A_t$ is obtained by integrating $v_\theta$ from $\tau=0$ to $1$, starting from Gaussian noise and using $K$ Euler steps. By explicitly localizing the task objects, $\mathcal{G}$ allows the action head to focus on learning how to manipulate them.

\subsection{Deployment}
\label{sec:method_deploy}

At episode start, $\mathcal{G}$ grounds object phrases parsed from language or directly receives 2D points or boxes from a human or VLM planner~\cite{molmo2}. CoTracker3~\cite{cotracker} then tracks points within the initial detection boxes and supplies them as prompts to $\mathcal{G}$, preserving object associations across frames. At each policy query, $\mathcal{G}$ constructs the current grounded observation, and the action head predicts a chunk conditioned on this observation and the instruction embedding using $K$ Euler steps. A prescribed number of actions is executed before querying again. For long-horizon and memory-dependent tasks, the planner uses the current image and episode memory to select new targets at sub-task transitions, reinitializing their trackers through new point prompts. This supports target selection informed by past observations while keeping the action policy unchanged across sub-tasks.

\section{Experiments}
\label{sec:experiments}
We design our experiments to answer four questions:
\textbf{(Q1)} Does a promptable 3D grounding foundation generalize to randomized scenes better than policies built on image encoders, point clouds, or VLMs?
\textbf{(Q2)} Does \method{} generalize beyond fixed training layouts to novel target locations and newly designated target objects?
\textbf{(Q3)} Does the unified task interface support deployment on real robots, both when a human specifies the target directly and when a VLM planner supplies target selections as points for long-horizon and memory-dependent manipulation?
\textbf{(Q4)} Do the gains come from restricting the observation to the task objects, and does the policy need both object appearance and metric 3D geometry?

\subsection{(Q1) Manipulation performance on RoboTwin 2.0}
\label{sec:exp_RoboTwin 2.0}

\begin{table}[t]
\caption{\textbf{Success rates (\%) on the 50 RoboTwin 2.0 tasks.}
Methods are grouped by backbone family and evaluated under the
official leaderboard protocol, which ranks entries by the average
of the clean (Easy) and randomized (Hard) settings regardless of
training scheme.
$\dagger$ denotes single-task training (one policy per task);
unmarked methods are co-trained on all 50 tasks. The best result in each column is bold.}
\label{tab:RoboTwin 2.0}
\centering
\footnotesize
\begin{tabular*}{\columnwidth}{@{\extracolsep{\fill}} l c c c @{}}
\toprule
Method & Easy & Hard & Avg. \\
\midrule

\multicolumn{4}{@{}l}{\textit{Action policies}} \\
DP$^\dagger$~\cite{dp}
    & 28.04 & 0.64  & 14.34 \\
ACT$^\dagger$~\cite{act}
    & 29.74 & 1.74  & 15.74 \\
RDT$^\dagger$~\cite{rdt}
    & 34.50 & 13.72 & 24.11 \\
DP3$^\dagger$~\cite{dp3}
    & 55.24 & 4.96  & 30.10 \\
\midrule

\multicolumn{4}{@{}l}{\textit{Vision-language-action models (VLAs)}} \\
starVLA~\cite{starvla}
    & 46.50 & 3.20  & 24.80 \\
$\pi_0^\dagger$~\cite{pi0}
    & 46.42 & 16.34 & 31.38 \\
GalaxeaVLA~\cite{galaxeavla}
    & 62.70 & 12.70 & 37.70 \\
Xiaomi Robotics-0~\cite{xiaomir0}
    & 62.90 & 18.20 & 40.50 \\
EventVLA~\cite{eventvla}
    & 65.60 & 15.70 & 40.60 \\
Abot-M0~\cite{abot}
    & 57.40 & 30.40 & 43.90 \\
X-VLA~\cite{xvla}
    & 68.00 & 20.90 & 44.50 \\
$\pi_{0.5}^{\dagger}$~\cite{pi05}
    & 64.00 & 25.90 & 44.95 \\
Spatial Forcing~\cite{sf}
    & 77.20 & 26.70 & 52.00 \\
\midrule

\multicolumn{4}{@{}l}{\textit{World-action models (WAMs)}} \\
AHA-WAM~\cite{ahawam}
    & 64.30 & 3.20  & 33.80 \\
FastWAM~\cite{fastwam}
    & \textbf{77.80} & 1.90 & 39.90 \\
X-WAM~\cite{xwam}
    & 70.00 & 25.80 & 47.90 \\
\midrule

\multicolumn{4}{@{}l}{\textit{Grounded action model (GAM)}} \\
\textbf{\method{} (ours)}$^\dagger$
    & 62.96 & \textbf{47.62} & \textbf{55.29} \\
\bottomrule
\end{tabular*}
\vspace{-20pt}
\end{table}

\textbf{Setup and baselines.} RoboTwin 2.0~\cite{robotwin2} comprises 50 bimanual manipulation tasks, each evaluated in a clean setting (Easy) and a randomized setting (Hard). Following the official single-task protocol, we train one \method{} policy per task on the 50 clean-scene demonstrations provided by the benchmark and evaluate 100 rollouts per task in both settings; no randomized scenes are seen during action policy training. Since the grounding backbone is pretrained on real images, we first adapt it to the rendered RoboTwin 2.0 domain by fine-tuning it alone, independently of any policy, on 2D boxes, 3D boxes, and depth exported directly from the simulator state, which requires no manual annotation. The adapted backbone is then frozen, and only the action head is trained per task. We compare against representative entries from the official leaderboard, grouped by backbone family as in Table~\ref{tab:RoboTwin 2.0}: action policies without a vision-language or video-generation backbone (DP, ACT, RDT, DP3), VLAs built on vision-language backbones ($\pi_0$, $\pi_{0.5}$, and the co-trained entries), and WAMs built on video-generation backbones. Baseline numbers are those reported by the RoboTwin 2.0 team, except for $\pi_{0.5}$, which we train and evaluate ourselves under the same single-task protocol.

\textbf{\method{} outperforms action policies, VLAs, and WAMs on average, and generalizes best to randomized scenes at test time.} As shown in Table~\ref{tab:RoboTwin 2.0}, \method{} achieves the highest average success among the listed methods (55.3\%) and, under the same single-task protocol, outperforms the strongest such baseline, $\pi_{0.5}$, by 10 points (55.3\% vs.\ 45.0\%). The margin is largest in the randomized setting, on which no method is trained: \method{} reaches 47.6\%, versus 30.4\% for the next best entry, and retains 76\% of its clean performance (63.0\% to 47.6\%). The strongest VLAs and WAMs, co-trained on all 50 tasks, reach higher clean success than \method{} (77.2\% for Spatial Forcing and 77.8\% for FastWAM), yet fall to 26.7\% and 1.9\% under randomization, and the best VLA and WAM entries on Hard remain at 30.4\% and 25.8\%. Their language and video pretraining transfers semantic and visual priors, but the policy still observes the whole scene, so randomized distractors and appearance enter the observation unchanged. The 3D policy DP3 is instructive in the same way: it consumes the scene point cloud directly and is competitive in clean scenes (55.2\%), yet drops to 5.0\% under randomization, showing that 3D input alone does not confer robustness. \method{} uses the same sensing but conditions only on the grounded task objects, so perturbations to distractors, background, and appearance largely do not enter its observation, and the backbone, adapted once to the rendered domain and frozen thereafter, localizes the task objects reliably in randomized scenes.

\subsection{(Q2) Generalization to test-time perturbations on LIBERO-PRO}
\label{sec:exp_robust}
\begin{table*}[t]
\centering
\small
\setlength{\tabcolsep}{4pt}
\caption{Success rates on LIBERO-PRO~\cite{liberopro} under four perturbations: object appearance and size (Obj), placement (Pos), instruction phrasing (Sem), and new targets or goals (Task). Avg.\ is the mean over 16 settings. Best and second-best results are bold and underlined, respectively, including ties.}
\label{tab:liberopro}
\begin{tabular}{l cccc cccc cccc cccc c}
\toprule
& \multicolumn{4}{c}{LIBERO-Spatial} & \multicolumn{4}{c}{LIBERO-Object} & \multicolumn{4}{c}{LIBERO-Goal} & \multicolumn{4}{c}{LIBERO-10} & \\
\cmidrule(lr){2-5} \cmidrule(lr){6-9} \cmidrule(lr){10-13} \cmidrule(lr){14-17}
Model & Obj & Pos & Sem & Task & Obj & Pos & Sem & Task & Obj & Pos & Sem & Task & Obj & Pos & Sem & Task & Avg. \\
\midrule
NORA~\cite{nora} & 0.92 & 0.00 & 0.91 & 0.00 & 0.86 & 0.00 & 0.92 & 0.00 & 0.58 & 0.00 & 0.88 & 0.00 & 0.46 & 0.00 & 0.74 & 0.00 & 0.39 \\
MolmoAct~\cite{molmoact} & 0.90 & 0.00 & 0.88 & 0.00 & 0.92 & 0.06 & 0.96 & 0.00 & 0.68 & 0.00 & 0.85 & 0.00 & 0.54 & 0.00 & 0.74 & 0.06 & 0.41 \\
$\pi_0$~\cite{pi0} & 0.95 & 0.00 & \textbf{0.97} & 0.00 & 0.94 & 0.00 & 0.90 & 0.00 & 0.94 & 0.00 & 0.93 & 0.00 & 0.79 & 0.00 & 0.82 & 0.00 & 0.45 \\
X-VLA~\cite{xvla} & \underline{0.97} & 0.00 & \underline{0.96} & 0.00 & 0.89 & 0.02 & 0.98 & \underline{0.08} & 0.68 & 0.01 & \textbf{0.98} & 0.09 & 0.62 & 0.00 & 0.95 & \textbf{0.10} & 0.46 \\
OpenVLA~\cite{openvla} & \underline{0.97} & 0.00 & \textbf{0.97} & 0.00 & \textbf{0.98} & 0.00 & 0.98 & 0.00 & \underline{0.96} & 0.00 & \textbf{0.98} & 0.00 & \underline{0.81} & 0.00 & \underline{0.96} & 0.00 & 0.48 \\
Flex-$\pi$~\cite{flexpi} & \textbf{0.99} & \underline{0.26} & \textbf{0.97} & 0.00 & \underline{0.97} & 0.05 & \textbf{1.00} & 0.00 & 0.76 & 0.11 & \underline{0.97} & \underline{0.10} & 0.77 & \underline{0.05} & \textbf{0.99} & \textbf{0.10} & 0.51 \\
$\pi_{0.5}$~\cite{pi05} & \underline{0.97} & 0.20 & \textbf{0.97} & \underline{0.01} & \textbf{0.98} & \underline{0.17} & 0.96 & 0.01 & \textbf{0.97} & \textbf{0.38} & \underline{0.97} & 0.00 & \textbf{0.92} & \textbf{0.08} & 0.93 & 0.01 & \underline{0.53} \\
\midrule
\method{} (ours) & 0.96 & \textbf{0.60} & 0.95 & \textbf{0.88} & 0.88 & \textbf{0.47} & \underline{0.99} & \textbf{0.50} & 0.69 & \underline{0.19} & \underline{0.97} & \textbf{0.16} & 0.50 & \underline{0.05} & 0.89 & \underline{0.08} & \textbf{0.61} \\
\bottomrule
\end{tabular}
\vspace{-10pt}
\end{table*}

\textbf{Setup and baselines.} LIBERO-PRO~\cite{liberopro} keeps the LIBERO training data, in which every task has a fixed object layout, and perturbs the tasks at test time along four axes: object appearance and size (Obj), object placement (Pos), instruction phrasing (Sem), and the target object or goal (Task). Obj and Sem leave the layout and goal intact, whereas Pos and Task change them; the latter two therefore test whether a policy grounds the instruction or replays a memorized trajectory. Following the LIBERO-PRO protocol, we train a single \method{} policy on the LIBERO demonstrations of all four suites and evaluate it under all four perturbations. We compare against $\pi_{0.5}$~\cite{pi05}, OpenVLA~\cite{openvla}, X-VLA~\cite{xvla}, $\pi_0$~\cite{pi0}, MolmoAct~\cite{molmoact}, and NORA~\cite{nora}, whose numbers are taken from the official LIBERO-PRO leaderboard, and Flex-$\pi$~\cite{flexpi}, which we evaluate ourselves using its released checkpoint.

\textbf{\method{} generalizes beyond fixed training layouts.} Table~\ref{tab:liberopro} shows that the baselines remain strong under Obj and Sem, the two perturbations that leave the layout and the goal intact, and collapse under Pos and Task, the two that change them; their success therefore tracks whether the trajectory memorized for the training layout is still valid, rather than whether the instruction is followed. \method{} is competitive on Sem and achieves particularly large gains on Spatial-Pos, Object-Pos, Spatial-Task, and Object-Task, because its policy is conditioned on grounding rather than on the raw scene; this yields the highest average across all 16 settings (0.61 vs.\ 0.53 for $\pi_{0.5}$). It trails under Obj on all four suites, most notably on LIBERO-Goal (0.69) and LIBERO-10 (0.50). \rev{Changes in object size shift the appropriate grasp locations. Although these geometric changes are represented in the object point clouds, the policy may struggle to adjust its grasping motions to object sizes not encountered during training.}

\textbf{Grounding enables spatial generalization to unseen target locations.} Under Pos, the target appears at a location never seen in training, yet the grounded observation still localizes it and the policy follows the detection: \method{} reaches 0.60 on Spatial and 0.47 on Object, two to three times the best baseline. The exception is Goal-Pos, where $\pi_{0.5}$ leads (0.38 vs.\ 0.19).

\textbf{Grounding enables task generalization to newly designated targets.} Task is the hardest setting because the instruction is replaced by one that designates a different object in the same scene, an object that appeared only as background in the original task; all objects and actions involved nonetheless appear in the training set. The memorized trajectory is therefore exactly wrong, and every baseline stays below 0.11 on all four suites. \method{} reaches 0.88 on Spatial, 0.50 on Object, and 0.16 on Goal, because grounding the new target changes the detection tokens and the policy acts on the newly grounded object at its actual location. In Task, as in Pos, the observation is defined by what is grounded rather than by what the instruction was paired with during training, so a change of target is reflected in the input instead of breaking a fixed visuomotor mapping.

\subsection{(Q3) Real-robot manipulation}
\label{sec:exp_real}

\begin{table}[t]
    \centering
\caption{Bimanual YAM results over 20 trials (5 per configuration). \method{} uses clicks or 2D boxes; $\pi_{0.5}$ uses language. Neither uses a planner or episode memory.}
    \label{tab:yam}
    \footnotesize
    \begin{tabular*}{\columnwidth}{@{\extracolsep{\fill}} l c c c @{}}
        \toprule
        Method & ID & OOD
        & Retention \\
        \midrule
        $\pi_{0.5}$~\cite{pi05}  & 19/20 & 4/20 & 21\% \\
        \textbf{\method{} (ours)} & 19/20 & \textbf{17/20} & \textbf{89\%} \\
        \bottomrule
    \end{tabular*}
    \vspace{-20pt}
\end{table}

\begin{figure*}[t]
    \centering
    \includegraphics[width=\textwidth]{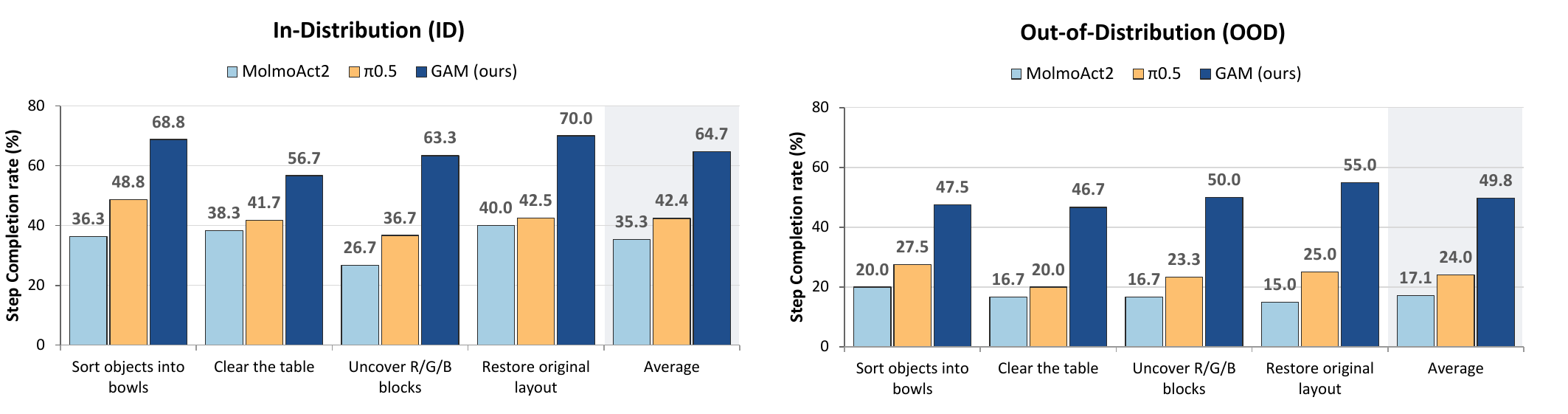}
    \vspace{-15pt}
    \caption{\textbf{Real-world results on the Franka.} Step completion rate (\%) on two long-horizon and two memory-dependent tasks in distribution (\textbf{left}) and out of distribution (\textbf{right}). \method{} uses a Molmo2 planner through its point interface; the baselines use their language interfaces without an external planner. Under distribution shift, \method{} retains 77\% of its ID performance, compared with 57\% for $\pi_{0.5}$ and 48\% for MolmoAct2.}
        \vspace{-3pt}
    \label{fig:franka}
\end{figure*}

\begin{figure*}[t]
    \centering
    \vspace{-10pt}
    \includegraphics[width=\textwidth]{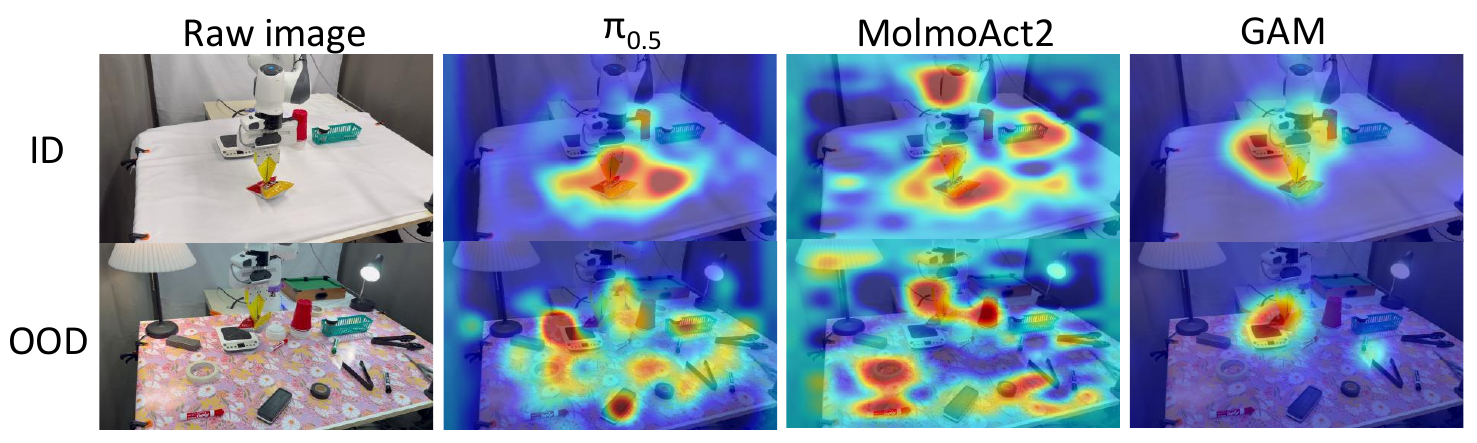}
    \caption{\textbf{Attention under distribution shift.} Action-head attention on \textit{Clear the table} in distribution (\textbf{top}) and out of distribution (\textbf{bottom}), min-max normalized per image. In these examples, the baselines' attention is more dispersed over the perturbed background and distractors, while \method{}'s remains concentrated around the selected targets.}
    \vspace{-20pt}
    \label{fig:attention}
\end{figure*}

\textbf{Setup and baselines.} We evaluate on two embodiments. On a
bimanual YAM, we evaluate a two-arm pick-and-place task with three
same-colored objects on each side, in which each arm places one
designated object into a shared container; four target combinations are
evaluated, designating two of the three objects on each side. The target
is specified directly, by a click or a 2D box for \method{} and through
the language interface for $\pi_{0.5}$~\cite{pi05}; neither system uses a
planner or episode memory. We collect 25 demonstrations per combination
and fine-tune both methods on the same data. We report success over 20
trials, five episodes for each of four combinations, in distribution and
under visual shift.

On a single-arm Franka, we evaluate four tasks with two to four sequential steps: two long-horizon tasks (\textit{Sort objects into three bowls}, \textit{Clear the table}) and two memory-dependent tasks (\textit{Uncover the R, G, B blocks in order}, \textit{Restore original layout}). A Molmo2 planner~\cite{molmo2} uses current observations and episode history to select objects via 2D points, while language specifies the operation for \method{}. All methods are fine-tuned on the same 50 demonstrations per task. We compare against $\pi_{0.5}$~\cite{pi05} and MolmoAct2~\cite{molmoact2} through their released language interfaces without an external planner. We report step completion rate (completed/total steps) over 20 ID and 10 OOD trials per task, with OOD changes to backgrounds, lighting, object instances, and colors, plus added distractors.

\textbf{The interface supports direct target specification on real hardware.} On the YAM, both methods achieve 19/20 successes in distribution. Under visual shift, with each system retaining its respective interface, \method{} achieves 17/20 successes, compared with 4/20 for $\pi_{0.5}$. The two methods retain 89\% and 21\% of their respective in-distribution success rates (Table~\ref{tab:yam}). These results demonstrate that \method{} supports direct visual target specification on real hardware and exhibits substantially stronger robustness under the tested visual changes.

\textbf{A planner can drive the same grounded observation through the point interface.} On the Franka, the \method{}--Molmo2 system achieves an average in-distribution step completion rate of 64.7\%, compared with 42.4\% for $\pi_{0.5}$ and 35.3\% for MolmoAct2 (Fig.~\ref{fig:franka}, left). The planner updates object selections as execution progresses and uses earlier observations to select targets in memory-dependent tasks. These selections pass directly to \method{} as points, while language specifies the operation and the action policy remains unchanged across sub-tasks. Under distribution shift, the system achieves 49.8\% step completion and retains 77\% of its in-distribution performance, compared with 24.0\% and 57\% retention for $\pi_{0.5}$, and 17.1\% and 48\% retention for MolmoAct2 (Fig.~\ref{fig:franka}, right). These results demonstrate that the point interface supports planner-driven long-horizon and memory-dependent execution under changing scene conditions. The qualitative attention maps in Fig.~\ref{fig:attention} are consistent with this robustness: in the shown examples, \method{}'s attention remains concentrated around selected targets, while the baselines' attention is more dispersed over the background and distractors.

\begin{table}[t]
    \centering
\caption{Success rates (\%) for observation ablations on 10 RoboTwin 2.0 tasks.}
    \label{tab:ablation}

    \footnotesize
    \setlength{\tabcolsep}{2pt}
    \renewcommand{\arraystretch}{1.15}

    \begin{tabular*}{\columnwidth}{
        @{\extracolsep{\fill}}
        >{\raggedright\arraybackslash}p{0.30\columnwidth}
        *{5}{c}
        @{}
    }
        \toprule
        \multirow{2}{*}{\textbf{Variant}}
        & \multicolumn{2}{c}{\textbf{Observation}}
        & \multicolumn{3}{c}{\textbf{Success}} \\
        \cmidrule(lr){2-3}
        \cmidrule(lr){4-6}
        & \textbf{Image}
        & \textbf{Detection}
        & \textbf{Easy}
        & \textbf{Hard}
        & \textbf{Avg.} \\
        \midrule

        \textbf{\method{} (full)}
        & grounded & grounded
        & \textbf{51.5} & \textbf{42.0} & \textbf{46.8} \\

        \midrule
        \multicolumn{6}{@{}l}{\textit{Removing object-centric filtering}} \\

        (a)~No image masking
        & full & grounded
        & 30.5 & 14.0 & 22.3 \\

        (b)~No point cropping
        & grounded & full
        & 18.5 & 7.0 & 12.8 \\

        (c)~Neither
        & full & full
        & 17.0 & 3.5 & 10.3 \\

        \midrule
        \multicolumn{6}{@{}l}{\textit{Removing one stream}} \\

        (d)~Detection only
        & -- & grounded
        & 19.5 & 12.5 & 16.0 \\

        (e)~Image only
        & grounded & --
        & 21.5 & 19.0 & 20.3 \\

        \bottomrule
    \end{tabular*}
    \vspace{-20pt}
\end{table}

\subsection{(Q4) Ablations}
\label{sec:exp_ablation}

\textbf{Object-centric filtering improves robustness.} We evaluate observation variants on 10 RoboTwin 2.0 tasks with 20 rollouts per task in both Easy and Hard, keeping the backbone, action head, training data, and schedule fixed. For whole-scene point clouds, we retain the same point budget. As shown in Table~\ref{tab:ablation}, removing image masking or point cropping reduces average success from 46.8\% to 22.3\% and 12.8\%, respectively, while removing both lowers it to 10.3\%. Without either filter, success falls from 51.5\% to 17.0\% on Easy and from 42.0\% to 3.5\% on Hard. The larger relative degradation on Hard supports the importance of excluding irrelevant scene content under randomization. Since all variants share the same frozen backbone, these gains arise from how its outputs are selected and represented rather than from a stronger perception backbone.

\textbf{Appearance and metric geometry provide complementary information.} Keeping only detection tokens or only image tokens yields average success rates of 16.0\% and 20.3\%, respectively, compared with 46.8\% for the full model; both variants retain the state-history token. Detection tokens encode metric object geometry, while image tokens preserve appearance cues around the task objects and robot arm. Neither stream alone matches their combination on Easy or Hard, supporting the value of jointly conditioning on visual and geometric information.
\section{Conclusion}
\label{sec:conclusion}
We presented Grounded Action Model (GAM), which connects task specification to action prediction through a pretrained promptable 3D detector. Object-centric visual and metric representations support robust manipulation under scene variation and generalization to relocated and newly designated targets, and its unified interface supports direct human target selection and planner-driven long-horizon and memory-dependent manipulation on real robots.

\textbf{Limitations and Future Work.} \rev{GAM depends on the quality and completeness of its grounded observations: grounding errors propagate to actions without recovery, and object-centric filtering can omit relevant context such as unselected obstacles. Improving this foundation calls for more diverse 3D grounding datasets and models that generalize across objects and scenes. The 3D bounding boxes used here are one possible form of grounding; alternatives such as dense 3D instance segmentation may provide finer-grained geometric supervision. We will explore these richer representations alongside grounding-aware failure recovery and adaptive context selection for safe execution.}


\bibliographystyle{IEEEtran}
\bibliography{references}

\end{document}